\documentclass[runningheads]{llncs}
\usepackage[T1]{fontenc}
\usepackage{graphicx,verbatim,booktabs}
\begin{document}
\title{Evaluation of Clinically Steerable Retinal Image Generation from Foundation Model Latent Spaces}

\author{Zuzanna A. Wakefield-Skórniewska\inst{1}\orcidID{0009-0000-9050-383X} \and
Bart\l omiej W. Papie\.z\inst{1}\orcidID{0000-0002-8432-2511}}
\authorrunning{Z. Wakefield-Skorniewska et al.}

\institute{Big Data Institute, Nuffield Department of Population Health, University of Oxford, United Kingdom\\
\email{zuzanna.skorniewska@ndph.ox.ac.uk} \\
\email{bartlomiej.papiez@bdi.ox.ac.uk}
}

\maketitle              
\begin{abstract}
Medical foundation models learn latent representations of clinically meaningful phenotypes, yet their ability to support controllable image generation remains largely unexplored. We evaluate four retinal foundation models within the representation tokenizer framework and examine whether demographic and clinical information encoded in latent representations from foundation models is preserved during synthetic image generation. We show that generated representations and images faithfully inherit phenotype information when evaluated within their originating foundation models, consistently outperforming conventional latent diffusion on multiple downstream prediction tasks. However, these gains largely disappear when evaluated using classifiers trained on real images, revealing a previously uncharacterised synthetic-to-real representation gap. These findings demonstrate that foundation-model latent spaces provide a powerful substrate for controllable retinal synthesis while highlighting the need to better align synthetic representations with real-image distributions.


\end{abstract}

\section{Introduction}
The recent advances in self-supervised learning (SSL) have driven the development of foundation models, which are trained on vast unlabelled datasets to learn transferable representations that support a wide range of downstream tasks.
In medical imaging, foundation models have emerged across modalities including brain Magnetic Resonance Imaging \cite{Mazher2026TowardsMRI,Wu2026BrainDINO:Learning}, chest radiography \cite{Tiu2022Expert-levelLearning,Ma2025ARadiography}, or retinal imaging \cite{Zhou2023AImages,Lee2026PRETI:Learning,Yu2024UrFound:Modeling,Silva-Rodriguez2025ASupervision}.
Within the field of oculomics, retinal foundation models (RFM) have shown remarkable performance in both retinal \cite{Zhou2023AImages,Lee2026PRETI:Learning,Silva-Rodriguez2025ASupervision,Yu2024UrFound:Modeling} and systemic health disease prediction \cite{Zhou2023AImages}, with learned representations often exhibiting stronger associations with clinical outcomes than conventional handcrafted retinal biomarkers \cite{Zhou2022AutoMorph:Pipeline}.

Beyond feature extraction, the learned representations may provide a structured latent space for controllable medical image synthesis. A central challenge in medical image generation is learning latent spaces that capture biologically  plausible variability while remaining steerable along clinically meaningful axes of variation. Conventional diffusion models \cite{Song2021DENOISINGMODELS}, however, are optimized for image reconstruction quality rather than the organization of diagnostically relevant features, and their latent spaces are therefore not explicitly structured around clinical phenotypes \cite{Skorniewska2026ExploringSynthesis}. We therefore hypothesize that diffusion within RFM latent spaces preserves clinically meaningful phenotypic structure more faithfully than conventional latent diffusion operating in generic VAE latent spaces.

To test this hypothesis, we evaluate four RFMs as latent spaces for controllable fundus image synthesis. We investigate whether demographic and clinical information encoded in RFM representations is retained in generated fundus images and compare this with conventional latent diffusion. We show that RFM latent spaces preserve clinically meaningful phenotypic information more effectively under foundation model-based evaluation. However, these gains diminish when synthetic images are evaluated using classifiers trained on real data, revealing a synthetic-to-real representation gap that highlights an important challenge for clinically reliable image synthesis.

\section{Methods}

\begin{figure}
    \centering
    \includegraphics[width=\linewidth]{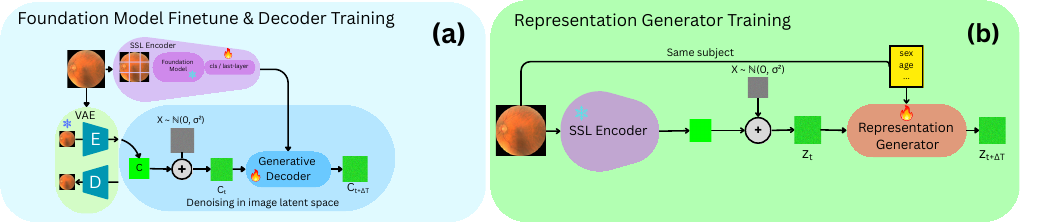}
    \caption{Overview of the proposed framework. (a) A generative decoder is trained to reconstruct VAE-compressed image latents conditioned on fine-tuned retinal foundation model representations. (b) A representation generator trained on outputs (representations) from the frozen foundation model with demographic and clinical conditioning for controllable generation.}
    \label{fig:diagram}
\end{figure}

\subsection{Representation Tokenizer}
To enable image synthesis, we employed the Representation Tokenizer (RepTok) framework \cite{Gui2025AdaptingGeneration}, a two-stage generative approach that performs diffusion directly in the representation space of a pretrained RFM using a lightweight, attention-free architecture. In the first training stage (see Fig.\ref{fig:diagram}a), a generative decoder is trained conditioned on feature representations from a fine-tuned foundation model (SSL Encoder), whose backbone remains frozen while only its final layer is optimised jointly with the decoder. The decoder, a DiT-B model \cite{Esser2024ScalingSynthesis}, receives this feature representation via token concatenation and is trained to denoise colour fundus images under a flow matching objective \cite{Lipman2022FlowModeling}. Prior to denoising, all images are compressed into spatially-reduced latents using a pretrained variational autoencoder (VAE) \cite{Rombach2022High-ResolutionModels}, following the original RepTok formulation. We retain the original cosine-similarity regularisation term, which encourages fine-tuned representations to remain close to their original embeddings while supporting pixel reconstruction learning.

In the second training stage (Fig.~\ref{fig:diagram}b), a 1D MLP-Mixer \cite{Tolstikhin2021MLP-Mixer:Vision} is trained via flow matching to generate CLS-token embeddings from subject-specific demographic and clinical metadata. Classifier-free guidance \cite{Ho2022Classifier-FreeGuidance} is incorporated by randomly dropping the conditioning inputs during the training, encouraging the model to learn both condition-specific and condition-invariant representations. At inference, these generated representations are then used to steer controllable image synthesis. 

\subsection{Retinal Foundation Models}
We evaluated RepTok using four RFMs spanning vision-only and vision-language models. Among the vision-only models, we employed \textbf{RETFound} \cite{Zhou2023AImages}, a ViT-L \cite{Dosovitskiy2020AnScale} encoder trained with masked image reconstruction \cite{He2021MaskedLearners} on 904,170 hospital-sourced fundus images, and \textbf{PRETI} \cite{Lee2026PRETI:Learning}, a ViT-B encoder trained with masked reconstruction conditioned on demographic variables (age and sex) on 1,017,549 fundus images. 
Among the vision-language models, we employed \textbf{FLAIR} \cite{Silva-Rodriguez2025ASupervision}, which combines a ResNet-50 \cite{He2015DeepRecognition} image encoder with BERT-encoded \cite{Devlin2018BERT:Understanding} condition descriptions under a CLIP-style contrastive training objective \cite{Radford2021LearningSupervision}, trained on 288,307 publicly available fundus images; and \textbf{URFound} \cite{Yu2024UrFound:Modeling}, a vision-language model using a ViT-B image encoder trained jointly on colour fundus and optical coherence tomography (OCT) images via masked image and language reconstruction, paired with BERT encoder for condition descriptions, trained on 180,000 publicly available retinal datasets. 
For image-generation fine-tuning, we optimise the CLS tokens of RETFound (1,024 parameters), URFound and PRETI (768 parameters each), and for FLAIR we instead optimise its own projection layer of 2048×512 (corresponding to approximately 1M parameters). We compare the four tested RepTok variants with a latent diffusion baseline that employs the same frozen VAE and a DiT-B as the diffusion backbone.

\subsection{Data}
We trained and evaluated all models on 256×256 colour fundus images from the UK Biobank \cite{Bycroft2018TheData}. 
For the first training stage (Fig.\ref{fig:diagram}a), we used 83,141 images, including both left- and right-eye images, focusing the task primarily on learning faithful image reconstruction. 
For the second training stage (Fig.\ref{fig:diagram}b), we used a subject-unique subset of 50,773 images, to learn the patient-specific representation conditioned on demographic and clinical variables. Conditioning variables included age, sex, body mass index (BMI), current hypertension status, as well as 3-year onsets of myocardial infarction (MI), stroke and chronic obstructive pulmonary disease (COPD).
For evaluation, we used a held-out test set of 11,780 unique subjects, with no subject overlap between the training and test data sets.

\section{Results}
\begin{figure}
    \centering
    \includegraphics[width=\linewidth]{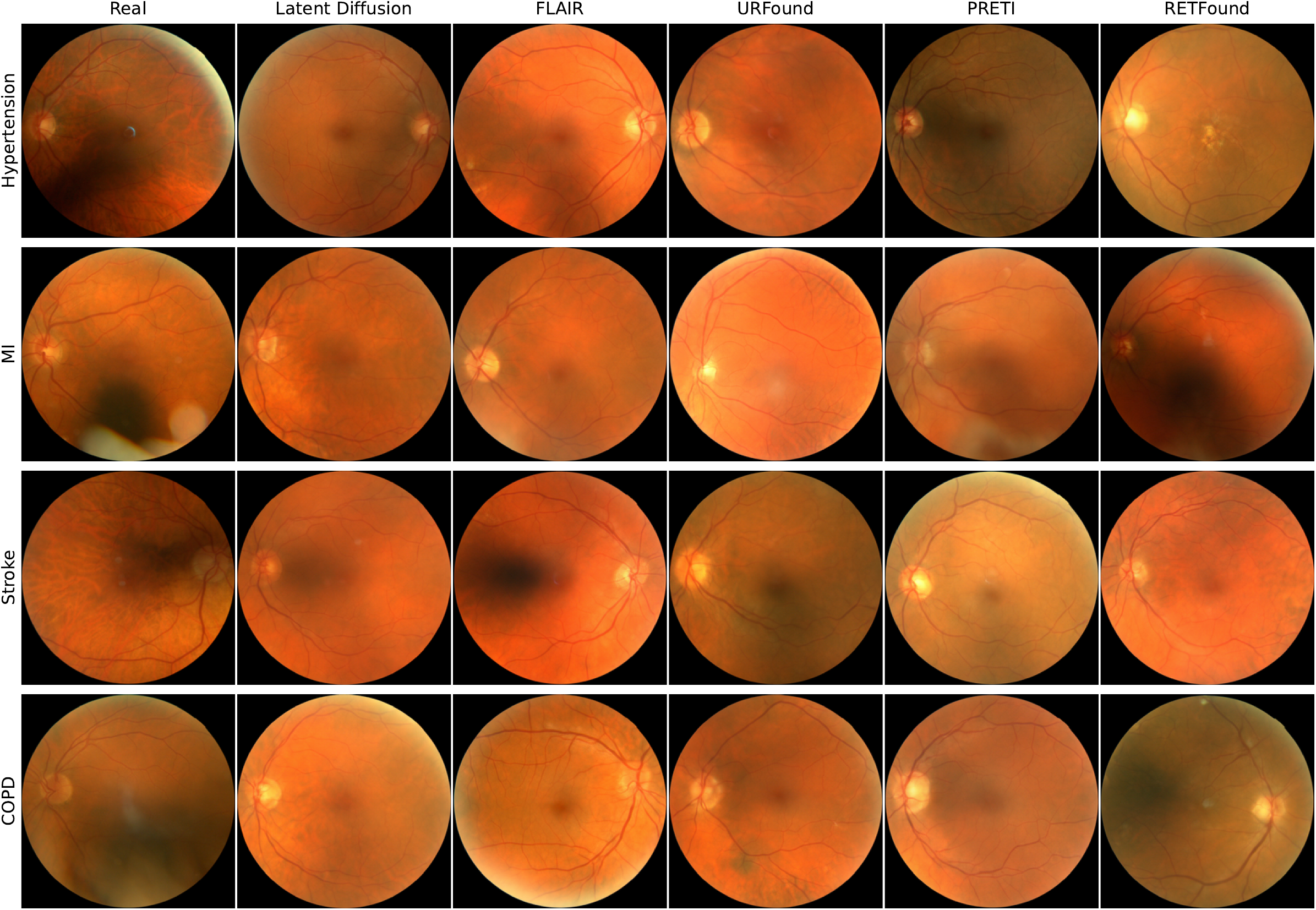}
    \caption{Real (first column) and synthetic fundus images generated by latent diffusion and RepTok conditioned on current hypertension and 3-year onset of chronic obstructive pulmonary disease (COPD), myocardial infarction (MI), and stroke.}
    \label{fig:samples}
\end{figure}

\subsection{Generation and Reconstruction Fidelity}

Table~\ref{tab:fid} summarizes reconstruction and generation quality using reconstruction Fréchet Inception Distance (rFID; images from generative decoder) and generation FID (gFID; images generated using tokens from representation generator). 
All RFMs achieved comparable reconstruction fidelity with low rFID values. However, generated image quality, measured by gFID, converged at values slightly higher than for the baseline latent diffusion model. 
Among the RepTok variants, FLAIR achieved the best generative performance (lowest gFID), potentially owing to its larger number of trainable parameters, but exhibited reduced reconstruction quality, as measured by peak signal-to-noise ratio (PSNR), structural similarity index (SSIM), and Learned Perceptual Image Patch Similarity (LPIPS), consistent with the reconstruction–generation trade-off reported in the original RepTok study.

Despite higher gFID values, qualitative inspection (Fig.~\ref{fig:samples}) shows that all RepTok models produce realistic retinal anatomy, including under rare clinical conditions. Similarly, Kernel Inception Distance (KID) computed on demographic subsets (Table~\ref{tab:kid_results}) yield comparable results for RepTok, latent diffusion, and real data, indicating that both approaches maintain high visual fidelity under conditional generation.

\begin{table}[t]
\centering
\caption{Reconstruction and generation quality metrics, evaluated on 10,000 real and 10,000 synthetic test images. The real baseline was computed using two independent sets of 10,000 real images.}
\label{tab:fid}
\begin{tabular}{llllll}
\hline
Generation Method & rFID ↓ & PSNR ↑ & SSIM ↑ & LPIPS ↓ & gFID ↓ \\ \hline
Real &  &  &  &  & 0.61 \\
Latent Diffusion &  & &  &  & 24.13 \\
RepTok (RETFound) & 3.29 & 30.37 & 0.86 & 0.18 & 34.79 \\
RepTok (PRETI) & 3.41 & 29.94 & 0.86 & 0.18 & 38.70 \\
RepTok (FLAIR) & 3.35 & 23.66 & 0.82 & 0.22 & 32.01 \\
RepTok (URFound) & 3.38 & 30.55 & 0.86 & 0.18 & 40.53 \\ \hline
\end{tabular}
\end{table}

\begin{table}[t]
\centering
\caption{Kernel Inception Distance (KID) reported as mean (std) between real and synthetic images for current hypertension and 3-year onset of myocardial infarction (MI), stroke, and chronic obstructive pulmonary disease (COPD). Lower scores indicate greater similarity between the synthetic and real image distributions.}
\resizebox{\linewidth}{!}{
\begin{tabular}{lcccccc}

\toprule
Characteristic & Real & Latent Diffusion & RETFound & PRETI & FLAIR & URFound \\
\midrule
\textbf{Hypertension} \\
\quad Yes (n=3942) & 0.000 (0.000) & 0.035 (0.004) & 0.024 (0.004) & 0.031 (0.004) & 0.027 (0.003) & 0.032 (0.004) \\
\textbf{MI (3y)} \\
\quad Yes (n=34) & 0.000 (0.000) & 0.026 (0.000) & 0.024 (0.000) & 0.023 (0.000) & 0.024 (0.000) & 0.029 (0.000) \\
\textbf{Stroke (3y)} \\
\quad Yes (n=15) & 0.001 (0.000) & 0.038 (0.000) & 0.038 (0.000) & 0.047 (0.000) & 0.014 (0.000) & 0.030 (0.000) \\
\textbf{COPD (3y)} \\
\quad Yes (n=18) & -0.004 (0.000) & 0.029 (0.000) & 0.036 (0.000) & 0.034 (0.000) & 0.038 (0.000) & 0.028 (0.000) \\
\bottomrule
\end{tabular}
}
\label{tab:kid_results}
\end{table}

\subsection{Evaluation on clinical and demographic phenotypes}

\begin{figure}
    \centering
    \includegraphics[width=\linewidth]{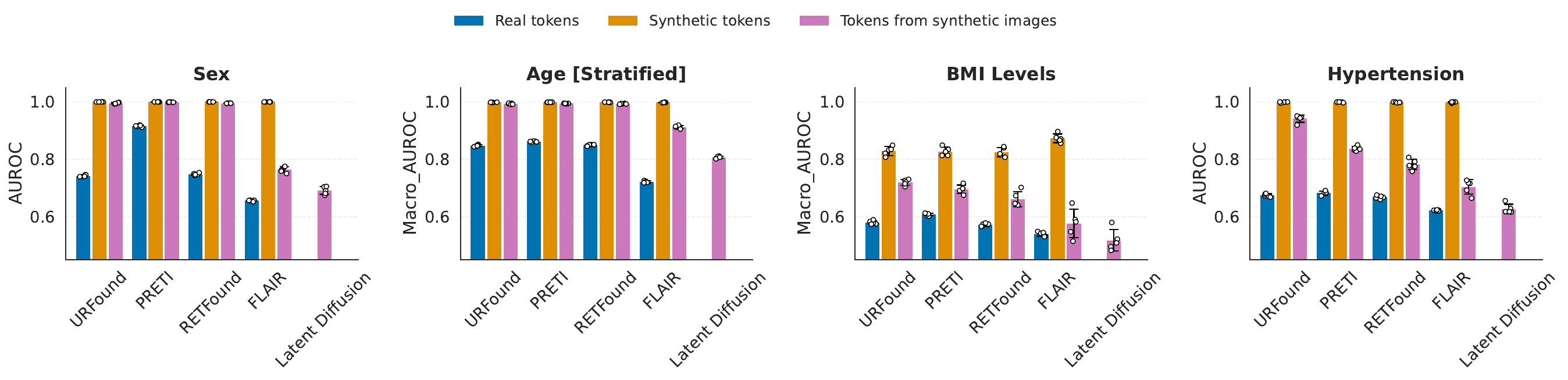}
    \caption{Prediction performance on demographic and hypertension tasks using foundation model representations from real images (blue), tokens from synthetic images (yellow), and synthetic images (purple). Overall, token from foundation model, tokens from synthetic images, and synthetic images all encode demographic information well, outperforming the baseline latent diffusion model.
}
    \label{fig:demo_auc}
\end{figure}

\begin{figure}
    \centering
    \includegraphics[width=\linewidth]{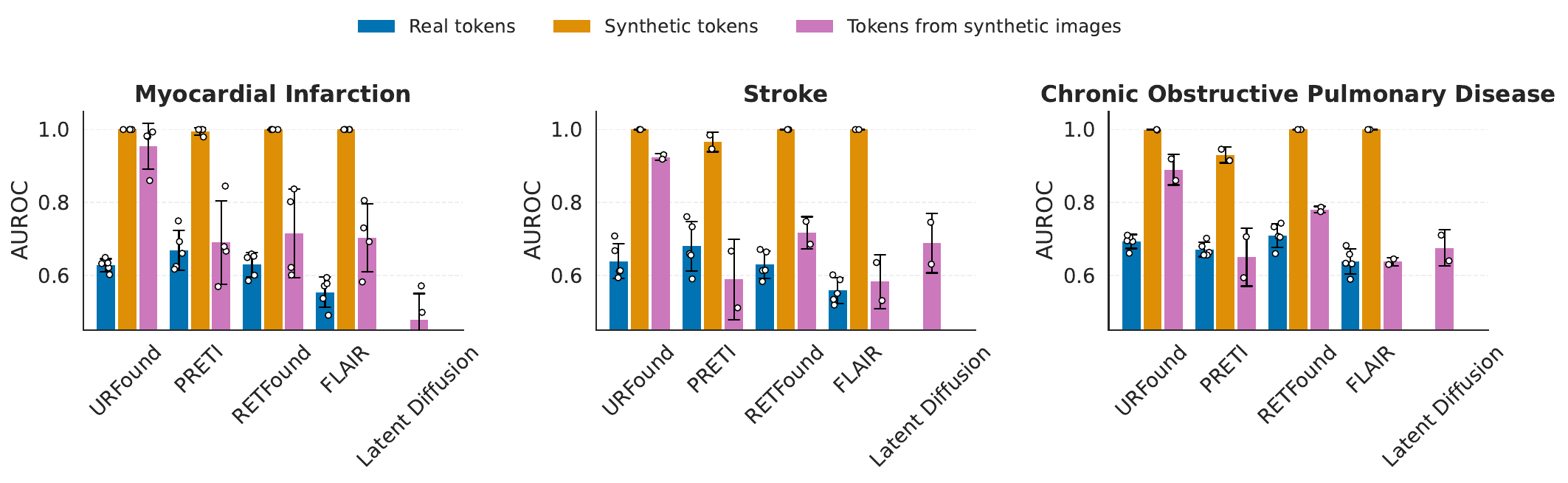}
    \caption{Prediction performance for 3 year onset of systemic conditions using foundation model representations from real images (blue), tokens from synthetic images (yellow), and synthetic images (purple).}
    \label{fig:condition_auc}
\end{figure}

We first assessed the phenotypic information encoded by each RFM using representations extracted from real images on a range of demographic and general health prediction tasks (Fig. \ref{fig:demo_auc}; blue bars) and 3-year systemic health onset prediction tasks (Fig. \ref{fig:condition_auc}; blue bars). All RFM achieved strong performance for sex, age, and hypertension prediction, with URFound and PRETI generally performing best, while BMI prediction remained more challenging across all models. For 3-year onset prediction, all models achieved AUROCs of between 0.6 and 0.7, although FLAIR consistently underperformed the remaining RFM, particularly for stroke and myocardial infarction. 

We next evaluated whether synthetic tokens from the representation generator preserved the conditioned phenotypic information (Figs. \ref{fig:demo_auc}, \ref{fig:condition_auc}; yellow bars). As shown, synthetic tokens achieved near-ceiling performance across all conditioned attributes, substantially outperforming representations extracted from real images. BMI remained the most challenging task but was still predicted considerably more accurately than from real-image representations. This strong performance is expected, as the representation generator is explicitly conditioned on the target demographic and clinical variables.

Finally, we evaluated whether phenotypic information was preserved after decoding synthetic representations into fundus images (Figs. \ref{fig:demo_auc}, \ref{fig:condition_auc}; purple bars). RepTok images achieved near-ceiling performance for sex and age prediction and generally outperformed real-image tokens for hypertension, BMI, and systemic disease onset prediction. URFound consistently achieved the strongest performance, particularly for disease onset prediction, while FLAIR and PRETI largely matched the performance of real-image representations. Compared with a latent diffusion baseline evaluated using URFound, RepTok achieved stronger myocardial infarction prediction performance across all models, with the diffusion model performing near chance. URFound was also the only model to consistently outperform latent diffusion across all remaining onset prediction tasks. For sex, age, and hypertension, latent diffusion achieved performance comparable to real URFound tokens, suggesting that the synthetic images preserve at least as much relevant information as the original data.

These results indicate that RFM latent spaces preserve clinically meaningful phenotypic information throughout both representation and image generation. However, the improved compatibility between synthetic images and their originating RFM also suggests that these representations may remain partially model-specific, motivating the external evaluation presented in the following section.

\subsection{External Validation}

\begin{table}[]
\centering
\caption{External validation of phenotype preservation in synthetic images using independently trained ResNet32 classifiers on real fundus images.}
\label{tab:external_validation}

\begin{tabular}{lccccc}
\hline
Imaging          & Age (R2)             & Sex (Acc.)                     & Hypertension (Acc.) &  &  \\ \hline
Real Data        & 0.623                & 0.725                         & 0.717              &  &  \\
Reconstructed    & 0.402                & 0.695                         & 0.698              &  &  \\ \hline
Latent Diffusion & 0.376                & \textbf{0.645}                & 0.717              &  &  \\
RETFound         & 0.265                & 0.576                         & 0.709              &  &  \\
PRETI            & 0.170                & 0.637                         & 0.714              &  &  \\
FLAIR            & 0.279                & 0.598                         & 0.698              &  &  \\
URFound          & \textbf{0.541}       & 0.587                         & \textbf{0.718}     &  &  \\ \hline
\end{tabular}%

\end{table}

We next evaluated phenotype preservation using ResNet32 classifiers trained exclusively on real images to predict age ($R^2$), sex, and hypertension. Incidence outcomes were omitted due to insufficient sample sizes for reliable classifier training.
As shown in Tab.~\ref{tab:external_validation}, the classifiers achieved strong performance on real test data (first row).
Reconstructed images (second row) yielded comparable performance for sex and hypertension, but age prediction dropped substantially, with an absolute reduction in explained variance of approximately 20$\%$ compared with the real images. This suggests that the classifier may exploit spurious confounding imaging artifacts present in real data that are subsequently removed during reconstruction.

Across synthetic images, the superiority of RepTok observed in the internal evaluation largely disappeared, with latent diffusion generally showing stronger conditioning adherence than all RepTok models except URFound. URFound slightly outperformed latent diffusion for hypertension and achieved substantially better age preservation (+16.5 percentage points in explained variance). Overall, these findings reveal a gap between internal and external evaluation: although synthetic images retain clinically relevant information when assessed within the synthetic domain, these gains are substantially attenuated when evaluated using classifiers trained on real data. This discrepancy may reflect both domain shift between synthetic and real images and differences in the features emphasized by foundation models and conventional ResNet classifiers. Notably, the marked reduction in age prediction performance observed for reconstructed real images suggests that external classifier performance is itself sensitive to image transformations and should therefore be interpreted with caution.

\section{Discussion}
In this work, we evaluated whether RFM latent spaces preserve clinically meaningful phenotypic information during controllable image generation. We show that synthetic images inherit demographic and clinical information encoded by RFM, enabling phenotype-preserving image synthesis across multiple clinically relevant attributes. However, these advantages are substantially attenuated under external evaluation with classifiers trained on real images, revealing a synthetic-to-real representation gap that limits downstream transferability. 

Among the evaluated RFM, URFound consistently achieved the strongest performance, outperforming latent diffusion in both the internal and external evaluations, whereas FLAIR generally performed worst. FLAIR's weaker performance may be partly explained by its CLIP-based training objective, which was also reported to underperform in the original RepTok study \cite{Gui2025AdaptingGeneration}. In contrast, URFound's strong performance relative to RETFound and PRETI suggests that multimodal pretraining may provide representations that are better suited to controllable image synthesis than vision-only pretraining. Interestingly, URFound achieved these results despite being trained on substantially fewer retinal images than RETFound or PRETI, indicating that pretraining strategy may be at least as important as dataset scale for this task.

Future work should focus on improving the alignment between synthetic and real image representations, enabling synthetic images to retain the clinically meaningful phenotypic information encoded by foundation models while remaining compatible with independently trained models. Bridging this synthetic-to-real representation gap would allow synthetic and real images to be used more interchangeably for downstream clinical tasks. Promising directions include learning representations that are both clinically informative and domain invariant, as well as developing generative objectives that better preserve predictive information across model families.
\\

{\small\noindent\textbf{Disclosure of Interests.}
The authors have no competing interests to declare that are relevant to the content of this article.} 
\\
{\small\noindent\textbf{Acknowledgement.}
This research has been conducted using data from UK Biobank with access provided through application 80521. ZAWS is supported by departmental funding from the Nuffield Department of Population Health, University of Oxford.

}
\bibliographystyle{splncs04}
\bibliography{references}

@article{Lipman2022FlowModeling,
    title = {{Flow Matching for Generative Modeling}},
    year = {2022},
    journal = {11th International Conference on Learning Representations, ICLR 2023},
    author = {Lipman, Yaron and Chen, Ricky T.Q. and Ben-Hamu, Heli and Nickel, Maximilian and Le, Matt},
    month = {10},
    publisher = {International Conference on Learning Representations, ICLR},
    url = {https://arxiv.org/pdf/2210.02747},
    arxivId = {2210.02747}
}

@article{Mazher2026TowardsMRI,
    title = {{Towards generalisable foundation models for brain MRI}},
    year = {2026},
    journal = {npj Imaging 2026},
    author = {Mazher, Moona and Parker, Geoff J. M. and Alexander, Daniel C.},
    month = {5},
    publisher = {Nature Publishing Group},
    url = {https://www.nature.com/articles/s44303-026-00176-5},
    doi = {10.1038/S44303-026-00176-5},
    issn = {2948-197X}
}

@article{Wu2026BrainDINO:Learning,
    title = {{BrainDINO: A Brain MRI Foundation Model for Generalizable Clinical Representation Learning}},
    year = {2026},
    author = {Wu, Yizhou and Wang, Shansong and Li, Yuheng and Safari, Mojtaba and Hu, Mingzhe and Chang, Chih-Wei and Veeraraghavan, Harini and Yang, Xiaofeng},
    month = {4},
    url = {https://arxiv.org/pdf/2604.27277},
    arxivId = {2604.27277}
}

@article{Tiu2022Expert-levelLearning,
    title = {{Expert-level detection of pathologies from unannotated chest X-ray images via self-supervised learning}},
    year = {2022},
    journal = {Nature Biomedical Engineering 2022 6:12},
    author = {Tiu, Ekin and Talius, Ellie and Patel, Pujan and Langlotz, Curtis P. and Ng, Andrew Y. and Rajpurkar, Pranav},
    number = {12},
    month = {9},
    pages = {1399--1406},
    volume = {6},
    publisher = {Nature Publishing Group},
    url = {https://www.nature.com/articles/s41551-022-00936-9},
    doi = {10.1038/s41551-022-00936-9},
    issn = {2157-846X},
    pmid = {36109605}
}

@article{Ma2025ARadiography,
    title = {{A fully open AI foundation model applied to chest radiography}},
    year = {2025},
    journal = {Nature 2025 643:8071},
    author = {Ma, Dong Ao and Pang, Jiaxuan and Gotway, Michael B. and Liang, Jianming},
    number = {8071},
    month = {6},
    pages = {488--498},
    volume = {643},
    publisher = {Nature Publishing Group},
    url = {https://www.nature.com/articles/s41586-025-09079-8},
    doi = {10.1038/s41586-025-09079-8},
    issn = {1476-4687},
    pmid = {40500447}
}

@article{Zhou2023AImages,
    title = {{A foundation model for generalizable disease detection from retinal images}},
    year = {2023},
    journal = {Nature 2023 622:7981},
    author = {Zhou, Yukun and Chia, Mark A. and Wagner, Siegfried K. and Ayhan, Murat S. and Williamson, Dominic J. and Struyven, Robbert R. and Liu, Timing and Xu, Moucheng and Lozano, Mateo G. and Woodward-Court, Peter and Kihara, Yuka and Allen, Naomi and Gallacher, John E.J. and Littlejohns, Thomas and Aslam, Tariq and Bishop, Paul and Black, Graeme and Sergouniotis, Panagiotis and Atan, Denize and Dick, Andrew D. and Williams, Cathy and Barman, Sarah and Barrett, Jenny H. and Mackie, Sarah and Braithwaite, Tasanee and Carare, Roxana O. and Ennis, Sarah and Gibson, Jane and Lotery, Andrew J. and Self, Jay and Chakravarthy, Usha and Hogg, Ruth E. and Paterson, Euan and Woodside, Jayne and Peto, Tunde and Mckay, Gareth and Mcguinness, Bernadette and Foster, Paul J. and Balaskas, Konstantinos and Khawaja, Anthony P. and Pontikos, Nikolas and Rahi, Jugnoo S. and Lascaratos, Gerassimos and Patel, Praveen J. and Chan, Michelle and Chua, Sharon Y.L. and Day, Alexander and Desai, Parul and Egan, Cathy and Fruttiger, Marcus and Garway-Heath, David F. and Hardcastle, Alison and Khaw, Sir Peng T. and Moore, Tony and Sivaprasad, Sobha and Strouthidis, Nicholas and Thomas, Dhanes and Tufail, Adnan and Viswanathan, Ananth C. and Dhillon, Bal and Macgillivray, Tom and Sudlow, Cathie and Vitart, Veronique and Doney, Alexander and Trucco, Emanuele and Guggeinheim, Jeremy A. and Morgan, James E. and Hammond, Chris J. and Williams, Katie and Hysi, Pirro and Harding, Simon P. and Zheng, Yalin and Luben, Robert and Luthert, Phil and Sun, Zihan and McKibbin, Martin and O’Sullivan, Eoin and Oram, Richard and Weedon, Mike and Owen, Chris G. and Rudnicka, Alicja R. and Sattar, Naveed and Steel, David and Stratton, Irene and Tapp, Robyn and Yates, Max M. and Petzold, Axel and Madhusudhan, Savita and Altmann, Andre and Lee, Aaron Y. and Topol, Eric J. and Denniston, Alastair K. and Alexander, Daniel C. and Keane, Pearse A.},
    number = {7981},
    month = {9},
    pages = {156--163},
    volume = {622},
    publisher = {Nature Publishing Group},
    url = {https://www.nature.com/articles/s41586-023-06555-x},
    doi = {10.1038/s41586-023-06555-x},
    issn = {1476-4687},
    pmid = {37704728}
}

@article{Lee2026PRETI:Learning,
    title = {{PRETI: Patient-Aware Retinal Foundation Model via Metadata-Guided Representation Learning}},
    year = {2026},
    journal = {Lecture Notes in Computer Science},
    author = {Lee, Yeonkyung and Han, Woojung and Jun, Youngjun and Kim, Hyeonmin and Cho, Jungkyung and Hwang, Seong Jae},
    pages = {523--533},
    volume = {15960 LNCS},
    publisher = {Springer Science and Business Media Deutschland GmbH},
    url = {https://link.springer.com/chapter/10.1007/978-3-032-04927-8\_50},
    isbn = {9783032049261},
    doi = {10.1007/978-3-032-04927-8\_50},
    issn = {16113349},
    arxivId = {2505.12233}
}

@article{Yu2024UrFound:Modeling,
    title = {{UrFound: Towards Universal Retinal Foundation Models via Knowledge-Guided Masked Modeling}},
    year = {2024},
    journal = {Lecture Notes in Computer Science (including subseries Lecture Notes in Artificial Intelligence and Lecture Notes in Bioinformatics) },
    author = {Yu, Kai and Zhou, Yang and Bai, Yang and Soh, Zhi Da and Xu, Xinxing and Goh, Rick Siow Mong and Cheng, Ching Yu and Liu, Yong},
    pages = {753--762},
    volume = {15012 LNCS},
    publisher = {Springer Science and Business Media Deutschland GmbH},
    url = {https://link.springer.com/chapter/10.1007/978-3-031-72390-2\_70},
    isbn = {9783031723896},
    doi = {10.1007/978-3-031-72390-2\_70},
    issn = {16113349},
    arxivId = {2408.05618}
}

@article{Silva-Rodriguez2025ASupervision,
    title = {{A Foundation Language-Image Model of the Retina (FLAIR): encoding expert knowledge in text supervision}},
    year = {2025},
    journal = {Medical Image Analysis},
    author = {Silva-Rodr{\'{i}}guez, Julio and Chakor, Hadi and Kobbi, Riadh and Dolz, Jose and Ben Ayed, Ismail},
    month = {1},
    pages = {103357},
    volume = {99},
    publisher = {Elsevier},
    url = {https://www.sciencedirect.com/science/article/pii/S1361841524002822},
    doi = {10.1016/J.MEDIA.2024.103357},
    issn = {1361-8415},
    pmid = {39418828},
    arxivId = {2308.07898}
}

@inproceedings{Song2021DENOISINGMODELS,
    title = {{DENOISING DIFFUSION IMPLICIT MODELS}},
    year = {2021},
    booktitle = {ICLR 2021 - 9th International Conference on Learning Representations},
    author = {Song, Jiaming and Meng, Chenlin and Ermon, Stefano}
}

@article{Gui2025AdaptingGeneration,
    title = {{Adapting Self-Supervised Representations as a Latent Space for Efficient Generation}},
    year = {2025},
    author = {Gui, Ming and Schusterbauer, Johannes and Phan, Timy and Krause, Felix and Susskind, Josh and Bautista, Miguel Angel and Ommer, Björn},
    month = {10},
    url = {https://arxiv.org/pdf/2510.14630},
    arxivId = {2510.14630}
}

@article{Esser2024ScalingSynthesis,
    title = {{Scaling Rectified Flow Transformers for High-Resolution Image Synthesis}},
    year = {2024},
    journal = {Proceedings of Machine Learning Research},
    author = {Esser, Patrick and Kulal, Sumith and Blattmann, Andreas and Entezari, Rahim and M{\"{u}}ller, Jonas and Saini, Harry and Levi, Yam and Lorenz, Dominik and Sauer, Axel and Boesel, Frederic and Podell, Dustin and Dockhorn, Tim and English, Zion and Rombach, Robin},
    month = {3},
    pages = {12606--12633},
    volume = {235},
    publisher = {ML Research Press},
    url = {https://arxiv.org/pdf/2403.03206},
    issn = {26403498},
    arxivId = {2403.03206}
}

@inproceedings{Rombach2022High-ResolutionModels,
    title = {{High-Resolution Image Synthesis with Latent Diffusion Models}},
    year = {2022},
    booktitle = {Proceedings of the IEEE Computer Society Conference on Computer Vision and Pattern Recognition},
    author = {Rombach, Robin and Blattmann, Andreas and Lorenz, Dominik and Esser, Patrick and Ommer, Bjorn},
    volume = {2022-June},
    doi = {10.1109/CVPR52688.2022.01042},
    issn = {10636919}
}

@article{Tolstikhin2021MLP-Mixer:Vision,
    title = {{MLP-Mixer: An all-MLP Architecture for Vision}},
    year = {2021},
    journal = {Advances in Neural Information Processing Systems},
    author = {Tolstikhin, Ilya and Houlsby, Neil and Kolesnikov, Alexander and Beyer, Lucas and Zhai, Xiaohua and Unterthiner, Thomas and Yung, Jessica and Steiner, Andreas and Keysers, Daniel and Uszkoreit, Jakob and Lucic, Mario and Dosovitskiy, Alexey},
    month = {5},
    pages = {24261--24272},
    volume = {29},
    publisher = {Neural information processing systems foundation},
    url = {https://arxiv.org/pdf/2105.01601},
    isbn = {9781713845393},
    issn = {10495258},
    arxivId = {2105.01601}
}

@article{Ho2022Classifier-FreeGuidance,
    title = {{Classifier-Free Diffusion Guidance}},
    year = {2022},
    author = {Ho, Jonathan and Salimans, Tim},
    month = {7},
    url = {https://arxiv.org/pdf/2207.12598},
    arxivId = {2207.12598}
}

@article{Dosovitskiy2020AnScale,
    title = {{An Image is Worth 16x16 Words: Transformers for Image Recognition at Scale}},
    year = {2020},
    journal = {ICLR 2021 - 9th International Conference on Learning Representations},
    author = {Dosovitskiy, Alexey and Beyer, Lucas and Kolesnikov, Alexander and Weissenborn, Dirk and Zhai, Xiaohua and Unterthiner, Thomas and Dehghani, Mostafa and Minderer, Matthias and Heigold, Georg and Gelly, Sylvain and Uszkoreit, Jakob and Houlsby, Neil},
    month = {10},
    publisher = {International Conference on Learning Representations, ICLR},
    url = {https://arxiv.org/pdf/2010.11929},
    arxivId = {2010.11929}
}

@article{He2021MaskedLearners,
    title = {{Masked Autoencoders Are Scalable Vision Learners}},
    year = {2021},
    journal = {Proceedings of the IEEE Computer Society Conference on Computer Vision and Pattern Recognition},
    author = {He, Kaiming and Chen, Xinlei and Xie, Saining and Li, Yanghao and Dollar, Piotr and Girshick, Ross},
    month = {11},
    pages = {15979--15988},
    volume = {2022-June},
    publisher = {IEEE Computer Society},
    url = {https://arxiv.org/pdf/2111.06377},
    isbn = {9781665469463},
    doi = {10.1109/CVPR52688.2022.01553},
    issn = {10636919},
    arxivId = {2111.06377}
}

@article{He2015DeepRecognition,
    title = {{Deep Residual Learning for Image Recognition}},
    year = {2015},
    journal = {Proceedings of the IEEE Computer Society Conference on Computer Vision and Pattern Recognition},
    author = {He, Kaiming and Zhang, Xiangyu and Ren, Shaoqing and Sun, Jian},
    month = {12},
    pages = {770--778},
    volume = {2016-December},
    publisher = {IEEE Computer Society},
    url = {https://arxiv.org/abs/1512.03385v1},
    isbn = {9781467388504},
    doi = {10.1109/CVPR.2016.90},
    issn = {10636919},
    arxivId = {1512.03385}
}

@article{Devlin2018BERT:Understanding,
    title = {{BERT: Pre-training of Deep Bidirectional Transformers for Language Understanding}},
    year = {2018},
    journal = {NAACL HLT 2019 - 2019 Conference of the North American Chapter of the Association for Computational Linguistics: Human Language Technologies - Proceedings of the Conference},
    author = {Devlin, Jacob and Chang, Ming Wei and Lee, Kenton and Toutanova, Kristina},
    month = {10},
    pages = {4171--4186},
    volume = {1},
    publisher = {Association for Computational Linguistics (ACL)},
    url = {https://arxiv.org/pdf/1810.04805},
    isbn = {9781950737130},
    arxivId = {1810.04805}
}

@article{Radford2021LearningSupervision,
    title = {{Learning Transferable Visual Models From Natural Language Supervision}},
    year = {2021},
    journal = {Proceedings of Machine Learning Research},
    author = {Radford, Alec and Kim, Jong Wook and Hallacy, Chris and Ramesh, Aditya and Goh, Gabriel and Agarwal, Sandhini and Sastry, Girish and Askell, Amanda and Mishkin, Pamela and Clark, Jack and Krueger, Gretchen and Sutskever, Ilya},
    month = {2},
    pages = {8748--8763},
    volume = {139},
    publisher = {ML Research Press},
    url = {https://arxiv.org/pdf/2103.00020},
    isbn = {9781713845065},
    issn = {26403498},
    arxivId = {2103.00020}
}

@article{Bycroft2018TheData,
    title = {{The UK Biobank resource with deep phenotyping and genomic data}},
    year = {2018},
    journal = {Nature 2018 562:7726},
    author = {Bycroft, Clare and Freeman, Colin and Petkova, Desislava and Band, Gavin and Elliott, Lloyd T. and Sharp, Kevin and Motyer, Allan and Vukcevic, Damjan and Delaneau, Olivier and O’Connell, Jared and Cortes, Adrian and Welsh, Samantha and Young, Alan and Effingham, Mark and McVean, Gil and Leslie, Stephen and Allen, Naomi and Donnelly, Peter and Marchini, Jonathan},
    number = {7726},
    month = {10},
    pages = {203--209},
    volume = {562},
    publisher = {Nature Publishing Group},
    url = {https://www.nature.com/articles/s41586-018-0579-z},
    doi = {10.1038/s41586-018-0579-z},
    issn = {1476-4687},
    pmid = {30305743}
}

@article{Zhou2022AutoMorph:Pipeline,
    title = {{AutoMorph: Automated Retinal Vascular Morphology Quantification Via a Deep Learning Pipeline}},
    year = {2022},
    journal = {Translational Vision Science {\&} Technology},
    author = {Zhou, Yukun and Wagner, Siegfried K. and Chia, Mark A. and Zhao, An and Woodward-Court, Peter and Xu, Moucheng and Struyven, Robbert and Alexander, Daniel C. and Keane, Pearse A.},
    number = {7},
    month = {7},
    pages = {12--12},
    volume = {11},
    publisher = {The Association for Research in Vision and Ophthalmology},
    url = {https://doi.org/10.1167/tvst.11.7.12},
    doi = {10.1167/TVST.11.7.12},
    issn = {2164-2591},
    pmid = {35833885}
}

@inproceedings{Skorniewska2026ExploringSynthesis,
  title={Exploring the Effectiveness of Deep Features from Domain-Specific Foundation Models in Retinal Image Synthesis},
  author={Sk{\'o}rniewska, Zuzanna and Papie{\.z}, Bart{\l}omiej W},
  booktitle={Annual Conference on Medical Image Understanding and Analysis},
  pages={293--305},
  year={2025},
  organization={Springer}
}

\end{document}